# Adjustable Method Based on Body Parts for Improving the Accuracy of 3D Reconstruction in Visually Important Body Parts from Silhouettes

Aref Hemati, Azam Bastanfard

**Abstract**

*This research proposes a novel adjustable algorithm for reconstructing 3D body shapes from front and side silhouettes. Most recent silhouette-based approaches use a deep neural network trained by silhouettes and key points to estimate the shape parameters but cannot accurately fit the model to the body contours and consequently are struggling to cover detailed body geometry, especially in the torso. In addition, in most of these cases, body parts have the same accuracy priority, making the optimization harder and avoiding reaching the optimum possible result in essential body parts, like the torso, which is visually important in most applications, such as virtual garment fitting. In the proposed method, we can adjust the expected accuracy for each body part based on our purpose by assigning coefficients for the distance of each body part between the projected 3D body and 2D silhouettes. To measure this distance, we first recognize the correspondent body parts using body segmentation in both views. Then, we align individual body parts by 2D rigid registration and match them using pairwise matching. The objective function tries to minimize the distance cost for the individual body parts in both views based on distances and coefficients by optimizing the statistical model parameters. We also handle the slight variation in the degree of arms and limbs by matching the pose. We evaluate the proposed method with synthetic body meshes from the normalized S-SCAPE. The result shows that the algorithm can more accurately reconstruct visually important body parts with high coefficients.*

## 1. Introduction

3D human body modeling is a significant problem in computer vision and one of the essential requirements in computer graphics, from creating a 3D avatar in video games and virtual reality to virtual clothing, special effects, biomedical engineering, and customized product design. Many papers have focused on modeling a 3D human body using silhouettes in the last decade. Silhouettes are robust image features that can prepare enough information cost-effectively to visualize a relatively high-accuracy digital human in the virtual environment.

Most silhouette-based approaches recently train CNN by silhouette images and key points [4],[21],[32],[31],[35]. Thus their shape parameters prediction is highly dependent on train data. Although these approaches are evolving, 3D body shapes resulting from these parameters cannot match the body contour in both front and side silhouettes. Therefore, some body parts with more complicated geometry do not form accurately. Moreover, most of these approaches focus on the accuracy of the whole body with equal priority for different body parts. In contrast, in most of the mentioned applications, the torso, including body part measurements such as shoulder, chest, waist, and pelvis, is more important than the others. However, some works focused on body parts. For example, Xi et al. [14] used PCA for each body segment or local deformation [25], But their result did not have enough accuracy.

In order to account for body parts in the silhouette-based reconstruction process, one of the main problems is recognizing body parts in both 3D body meshes and 2D images and matching them. About 3D body models, most statistical models defined body mesh segmentation, in which body parts have the same vertices and faces in datasets. But recognizing body parts in 2D images requires an additional stage known as body segmentation in images. How can we find body parts without using this time-consuming stage? This problem is the first challenge and would be more vital when we do not limit the person pose strictly in photography assumptions.

The second challenge is to align and match the contours of the projected 3D body and 2D image to find the accurate distances between each part. The third issue is to define the objective function in terms of different parameters, including body part distances and coefficients, the weight of the front, side view, and height. The fourth challenge is how the body parameters optimization algorithm can optimally minimize the cost function by generating new models with the best statistical model parameters.

These four challenges we faced were specific problems related to our proposed method. However, there are some challenges that are common in silhouette-based reconstruction methods. For example, due to lens distortion, 2D photos

Aref Hemati, Azam Bastanfard
Department of Computer and Information Technology, Engineering, Qazvin Branch, Islamic Azad University, Qazvin, Iran
aref.hemati@qiau.ac.ir, bastanfard@kiau.ac.ir

from the front and side of the subject before processing should be undistorted. Silhouette extraction is the next challenge. To extract silhouettes from images three approaches can be use: background subtraction [41],[56],[24], semantic segmentation [42], or multi-view segmentation (visual hull) [43]. We simplify this process by defining a uniform color background assumption and extracting silhouettes from the background in both images using background subtraction [41]. To extract the silhouette boundary, some works [7],[8] used the Canny edge detector [51]. But we utilize the Moore-neighbor tracing algorithm [17],[18].

The statistical models are another common issue in the area of 3D body reconstruction and play a vital role in the results of approaches. The SCAPE [6] and, recently, SMPL [23] are the most popular ones between statistical models. There are different criteria for choosing the best statistical model, such as parameters required for input and quality of output 3D meshes, especially in various poses. A statistical model called normalized S-SCAPE [1] is used as a generative 3D mesh regression. The S-SCAPE model defines 100 parameters to be applied on 100 eigenvectors corresponding to 100 top eigenvalues to prepare a 3D shape. This model also used 30 angle parameters to create a pose for a 3D shape. However, in this work, we only exploit 20 parameters for the shape and four for the pose. It means 20 eigenvectors corresponding to 20 top eigenvalues are used for modeling a shape. In addition, these four pose parameters support a slight variation in the angles of the arms and legs of the subject and enable the person to not care about the accurate pose.

**Contribution.** In summary, this paper has the following contributions:

- We present a customizable method that enables us to significantly increase the accuracy of highly important body parts in reconstruction based on our purpose and application (including individual body parts, front and side views, and height).
- We introduce a novel method to recognize, align and match the correspondent body part boundaries between images and projected 3D meshes based on 3D body segmentation in meshes.
- A slight variation in the angles of the arms and legs of the person is handled by matching the pose, so the person does not restrict to having a fixed pose.

In section three, we discuss the proposed method, including the photography assumptions, silhouette boundary extraction method, S-SCAPE model and its parameters, aligning and matching each body part, defining coefficients, cost function, and body parameter optimization algorithm. We present the experimental result in section four and evaluate the result of the methods in section five. We discuss some issues in section six and conclude the work in section seven.

## 2. Related Works

Although many methods were presented for 3D human body modeling from images, they all follow three essential principles for constructing the 3D human body from images.

**Matching 3D model and 2D image:** Matching and measuring the distance between the 3D model and the 2D image is the first principle, and different approaches were proposed for this aim. In some works [7],[8], 60 maximum or minimum local points on the boundary of the silhouette are extracted as key points and used to compare 2D images and projected 3D models. In similar work [40], 2-15 feature points on the silhouettes were used to control matching errors between the model and image. Methods such as shoulder point detection (SPD) [60] can be utilized for extracting these points. The approach used by Zhang et al. [9] is mapping the measures, such as breadths of the shoulder, waist, and hip of a silhouette and template model. In work by Pavlakos et al. [34], 2D pose data using 2D articulated human pose estimation methods [38],[39] were extracted and utilized as 2D key points along with masks to learn a mapping to SMPL[23] model parameter. Kolotouros et al. [35] utilized iterative fitting on 2D joints to train a CNN. Bogo et al. [37] applied the DeepCut CNN [36] to predict 2D body joints in 2D images and defined an objective function for minimizing the error between the projected 3D model joints and detected 2D joints. Yan et al. [54] used regression to learn a mapping from 2D image to body measurement. This mapping allows them to predict measurements from images and reconstruct the body using the measures.

**Model fitting techniques:** The second principle is the way used for model fitting. Two model fitting techniques include mesh fitting (template deformation) and mesh regression using statistical models. In the mesh fitting technique, an initial template mesh is deformed using deformation techniques like free-form deformation (FFD) [57] or Laplacian-based deformation [46],[33]. An energy minimization function using a deformation technique tries to minimize the distance error of correspondent 60 feature points [8],[9] or 12 to 15 feature points [40] to reach the target 3D body mesh.

Nonetheless, many 3D human body datasets were released in the last decade. Civilian American and European Surface Anthropometry Resource (CAESAR) [5] dataset is a collection of thousands of range scans and 3D anthropometric landmarks. This dataset was collected from 18 to 65 age volunteers in Europe and North America. Hasler et al. [29] introduced a statistical model trained on ScanDB [30] that includes approximately 550 full-body 3D laser scans of 114 subjects. Size-UK [27] is a large public 3D human body dataset measured in a stationary position using stationary scanners. SCAPE [6] prepared a statistical model that supports both pose and shape deformation for the first time. FAUST [28] presented 300 real human body scans (10 subjects, 30 poses) with automatically computed dense ground-truth correspondences. It developed a novel scan-to-scan correspondence and texture-based registration technique to improve fitting. Allen et al. [3] used a subset of the meshes in the CAESAR dataset, 250 human body models consisting of 125 male and 125 female scans. They fit a template to these body models and made a body shape variation by applying PCA to them. In addition, Some MoCap datasets like Human3.6M [49] and HumanEva [50], which are video datasets with ground truth motion, were released. These datasets are used to calculate the precise ground truth 2D and 3D keypoint locations and contribute to pose estimation. This variety of datasets inclined most researchers to use the mesh regression technique.

Two powerful statistical body shape models created using PCA on datasets are SCAPE [6] and SMPL [23], which present low-dimensional parameter space to generate 3D human body shape and pose. To generate 3D mesh using the SCAPE [6], some works [44],[45] estimated SCAPE model parameters from silhouettes and key points. SMPL [23], as a parametric statistical body shape model, is one of the most popular statistical models and has been recently utilized in several works. In recent years most works that used SMPL computed its parameters using deep neural networks [34]. Kolotouros et al. [47] used a graph CNN to deform the initial mesh generated by SMPL. Li et al. [22] exploited SMPL to estimate the initial model and then warped geometry using 2D non-rigid registration to reach the final result. As we already mentioned, Kolotouros et al. [35] train a CNN using iterative fitting on 2D joints and fit the SMPL model on 2D joints. Bogo et al. [37] present SMPLify to estimate 3D body shape and pose in single images. In this work, to predict 2D body joint locations, a CNN method called DeepCut [36] was used, and then SMPL was fitted to predicted 2D joints.

SMPL is more visually symmetric and has vertex-based transformations in comparison with triangle-based transformations, which are used in SCAPE [6]. The S-SCAPE model is a statistical model proposed by Jain et al. [48], a simplified SCAPE model with better reconstruction time. The S-SCAPE results from PCA on the vertex coordinates and uses LBS approaches to change the pose. Yang et al. [20] used the S-SCAPE model for the human body and shape estimation under motion in 3D input sequences. Pishchulin et al. [53] utilized the S-SCAPE model to generate a 3D model with different shape and pose parameters. Then rendered these models from random viewpoints and added different background edges to generate non-photorealistic images and enrich training data with complementary shapes instead of collecting more training data. Older works like Boisvert et al. [12] defined a linear mapping from the 2D PCA space (silhouette) into 3D PCA space (3D data) using a Gaussian process latent variable model [55]. We use the normalized S-SCAPE model [1] as a statistical model. This model applied different data pre-processing to improve body shape models and statistical shape space qualities.

**Body parameters estimation algorithm:** The third principle in image-based body reconstruction is the method utilized for estimating 3D human body parameters to reach the best output mesh from the input image. The algorithms used for estimation depend on the model fitting method. In works that use mesh fitting, an objective or mapping function controls the deformation process. For example, Lin et al. [8] and Seo et al. [40] defined a map function that controlled free-form deformation (FFD) [57] error to fit the mesh to control points of silhouettes. Similar work by Zhang et al. [9] defined an objective function to control the Laplacian-based deformation process [46],[33]. Boisvert et al. [12] used both mesh fitting and regression techniques. After generating a 3D model using a statistical model, an energy function minimizes mesh deformation locally.

In mesh regression-based works, the algorithm used for estimation tries to predict the statistical shape and pose parameters to increase the similarity of the 3D model to the target. Xi et al. [14] mapped a 2D contour PCA space into a 3D PCA space using linear regression to calculate a relationship matrix used for predicting new data. Dibra et al., in their first work [15], trained a Canonical Correlation Analysis (CCA) regression forests semi-suppervisely by features extracted from large silhouette data and then used it to predict the body parameters. In the next work, Dibra et al. [21], for the first time, used CNNs to learn a global mapping from silhouettes to shape parameters. In the following, Dibra et al. [32] first trained a CNN with 3D shape descriptors (Heat Kernel Signature, HKS) [58], which are invariance to isometric deformations and maximize intrahuman-class variation to generate a richer body shape representation space. Then by learning a CNN from silhouette, they mapped 2D silhouette images to this new shape representation space. Smith et al. [4], with a multi-task learning approach, trained a convolutional model from front and side view silhouettes and predicted not only shape and pose parameters for SMPL but also 3D joints and mesh volume. In General, most recent works [30],[37],[35] utilized convolutional neural networks to train a model from inputs including silhouettes and 2D key points, and the trained model can estimate the shape parameters of the statistical model for every new image.

## 3. Proposed Method

In this section, we describe the technical details of the proposed method in different subsections.

### 3.1. Feature extraction

Silhouettes are a rich source of data in images, including the local maximum and minimum points and the curves. The more features achieved from silhouettes, the more accurate the 3D body can make. In this work, the whole boundary of silhouettes is used. But before extracting the boundary, some steps should be taken.

**Photography assumptions:** Some assumptions are considered to reduce the complexity of preparing the images. The background has a uniform color, and two photos are taken, one from the front and one from the side view. The subject has a standard posture like 'A' with straight limbs and arms apart from the torso. However, the method handles a slight variation in the angles of arms and limbs. In photos from the side view, hands should stick to the body so that no silhouette of hands appears in the side view silhouettes. No assumptions about the distance from the camera are made. However, two images should be taken directly from the front and side at the same height above the ground and cover the whole person's body. The image resolution acquired by the camera should be at least 640*480 pixels. However, super-resolution approaches, such as the method proposed by Bastanfard et al. [59], can improve the quality of low-resolution images.

In order to remove radial lens distortion of images acquired from the digital camera, the camera calibration toolbox in Matlab [16] is used. This toolbox extracts the internal and external parameters of the used digital camera by several chessboard images from different views, as shown in Fig. 1. Radial (lens) distortion is one of these internal parameters. The value of radial distortion obtained from the used camera is depicted in Table 1.

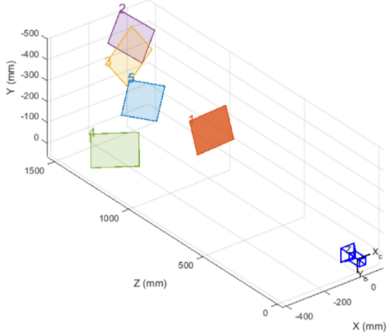

**Fig. 1:** Camera Calibrator toolbox calculates intrinsic and extrinsic parameters using serval images taken by the digital camera.

Table 1: Radial distortion of the used camera.

| Radial distortion |
| --- |
| [-0.1124 +/- 0.0414    0.5875 +/- 0.2488] |

The toolbox adopts a pinhole model to correct the radial lens distortion [19] in images. Before extracting silhouettes, every front and side photo from the subject should be undistorted.

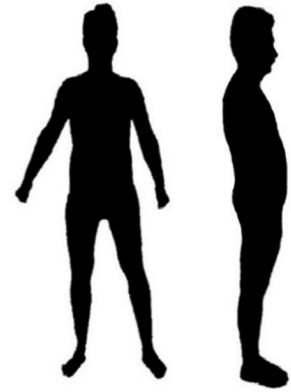

**Fig. 2:** Front and side silhouette of the subject.

**Pre-processing images:** Due to photography assumptions, detecting the subject in front of a uniform background in images is simple. We use background subtraction [55] to extract the silhouettes, as shown in Fig. 2.

**Silhouette boundary extraction:** In order to extract silhouette boundaries from both views, the Moore-neighbor tracing algorithm [17],[18] is used. Moore-neighbor is a boundary tracing method that works only for the closed region. It traced the boundary of the silhouette with a contiguous line.

Let P denote the current boundary pixel. The algorithm first locates a start pixel. Then it backtracks, marks the start pixel as a boundary pixel, and calls it P (current boundary pixel). Then around P in a clockwise direction, each pixel in its Moore neighborhood is visited until it hits a new boundary pixel. Again it backtracks and marks the newly founded pixel as a boundary pixel and calls it p. This operation repeats until it arrives back at the start pixel from the same direction as it was initially entered. In Fig. 3, pixels 1-8 are extracted boundary of shape using the Moore-Neighbor algorithm.

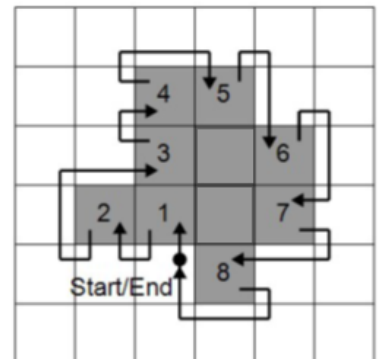

**Fig. 3:** An example of Moore-Neighbor boundary tracing found 1-8 pixels as a boundary [52] .

### 3.2. Model fitting

Model fitting is called the technique that is used to find the target 3D mesh from a given input. Inputs can be 3D scans, 2D or 3D key points, silhouettes, or point clouds. In this work, silhouettes are used as input. There are two approaches for model fitting: mesh fitting (registration or deformation) and mesh regression using statistical models. The mesh regression method utilizes template meshes because the number of vertices is fixed, and corresponding vertices have the same structure and semantics for all meshes in datasets. Therefore, obtaining measurements for one mesh can be used for other meshes in datasets. In this work, we use a statistical model for mesh regression.

**The statistical model:** Statistical models are sets of human bodies in different shapes and pose demonstrated by principle component (PCs). Principle component analysis (PCA) describes different pose and shape variations. PCA reduces high-dimension 3D meshes by finding the pose and shape's principal components which have the maximal variance in the dataset. These PCs are used to generate new meshes [2],[3],[4] from a pose-shape parameter space. There are plenty of 3D scans Datasets. The CAESAR [5] has over 4500 3D scans of American and European subjects, and SCAPE [6] is the first statistical model for both pose and shape deformations. Both are mainly used to build statistical models. This paper uses a simplified normalized version of the SCAPE model called S-SCAPE [1] as a statistical model.

**The S-SCAPE model:** Our work exploits normalized S-SCAPE as a generative model. It has a default A-pose as a standard pose. This model uses linear space learned using principal component analysis (PCA) to generate various shapes. In addition, a linear blend skinning (LBS) model is used to generate different poses. Therefore, two steps are taken to reconstruct a model. First, shape parameters are used to calculate a mesh with default fixed pose parameters. Then a linear blend skinning (LBS) model uses joint angle parameters to repose shape to the final mesh. In the following, we describe the S-SCAPE model introduced by Jain et al. [48] and well expressed by Yang et al. [20]. The expression $s(\beta, \Theta) \in R^{3N_v}$ indicates vector of vertex coordinates for shape $\beta$ and posture $\Theta$ and $N_v$ denotes the number of vertices. Let $\tilde{s}(\beta, \Theta) \in R^{4N_v}$ denote the vector, including corresponding homogeneous vertex coordinates. S-SCAPE model applies the PCA model for fixed posture $\Theta_0$ as the following expression:

$$\tilde{s}(\beta, \Theta_0) = \tilde{A}\beta + \tilde{\mu} \qquad (1)$$

Where $\tilde{A}$ indicates eigenvectors matrix founded by PCA and $\tilde{\mu}$ is the mean body shape coordinates. Then S-SCAPE model applies the posture using LBS that causes changes in shape pose with fixed shape size as the following expression:

$$s_i(\beta_0, \Theta) = \sum_{j=1}^{N_b} \omega_{ij} T_j(\Theta)\tilde{s}_i(\beta_0, \Theta_0) \qquad (2)$$

Where $s_i$ and $\tilde{s}_i$ Indicates homogenous coordinate vector of i-th vertex of s, number of bones used for linear blend skinning (LBS) is $N_b$ and transformation matrix used for the j-th bone is $T_j(\Theta)$. The rigging weight binding the i-th vertex to the j-th bone is $\omega_{ij}$. Equations (1) and Equations (2) combined as:

$$s(\beta_0, \Theta) = T(\Theta)\tilde{A}\beta + T(\Theta)\tilde{\mu} \qquad (3)$$

Where $T(\Theta) \in R^{3N_v \times 4N_v}$ is a sparse matrix containing per-vertex transformations.

**Statistical model parameters:** Based on the previous section, the S-SCAPE model needs three input types to generate a 3D model. The first one is a mesh template that consists of $N_v$=6449 vertices. The second input is shape parameters $\beta$, which are coefficients that multiply by the eigenvectors of PCs. This work uses the 20 eigenvectors corresponding to the first 20 eigenvalues. For example, the eigenvector corresponding to the highest eigenvalue is related to the attribute with maximal variance in the dataset. The height of body meshes has the maximal variance in the dataset. Therefore to generate a tall body mesh,$+3\sigma$ as a coefficient for this eigenvector can be applied. Similarly, to generate a short body mesh, $-3\sigma$ coefficient is used.

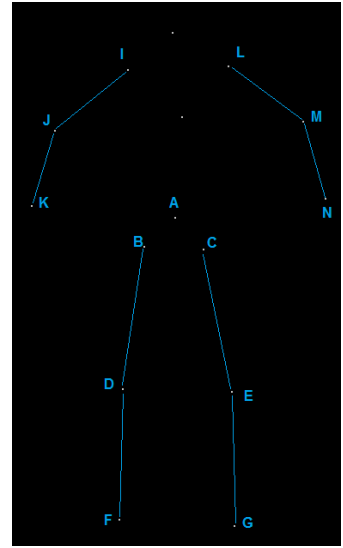

**Fig. 4:** 15 joints defined in S-SCAPE model. These joints include 30 joint angle parameters with movements on the X, Y, or Z-axis.

There are 20 coefficients that apply to 20 eigenvectors corresponding to the first 20 eigenvalues of PCs. Therefore

20 shape parameters defined for generating a mesh body shape. The third input for the S-SCAPE model is joint angle parameters that LBS uses to repose the default shape pose to the desired one. There are 30 joint angle parameters for 15 joints in template mesh, as shown in Fig. 4.

Some of these 30 joint angle parameters include: left hip rotation x, left hip rotation y, left knee, left foot, right hip rotation x, right hip rotation y, right knee, right foot, left humerus x (shoulder), left humerus y, left humerus z, right humerus x, right humerus y, right humerus z, left hip z, and right hip z. In assumptions, the subject has a standard posture like 'A' with straight limbs and arms apart from the torso. However, a slight variation (30 DOF) in the angles of arms and limbs would be inevitable. These changes are observed in four joints, including I, L, F, and G, in Fig. 4. The movement in these four joints for the subject in 'A' posture is limited only on the Y axis. Therefore, among 30 joint angle parameters, four joint angles related to I, L, F, and G joints include left hip rotation y, right hip rotation y, right humerus y, and right humerus y. Therefore, four pose parameters are used as the third input requirement of the S-SCAPE model to handle slight pose variation. Both shape and pose parameters are combined, resulting in 24 statistical model parameters, which are improved in the optimization process.

**3D human body projection:** The generated 3D human bodies in each iteration of the optimization algorithm using the statistical model, S-SCAPE, need to be compared with the front and side silhouettes of the subject. To compare 3D models with the person's silhouettes, they are projected from 3D to 2D space. Therefore, a projection matrix should be applied to the 3D model. Three matrices are multiplied to each 3D model to obtain the 2D projection of the model according to the following expressions:

$$T_x(R_x,v,p) = R_x \times v + p \times I \quad (4)$$
$$T_y(R_y,v,p) = R_y \times v + p \times I \quad (5)$$
$$T_z(R_z,v,p) = R_z \times v + p \times I \quad (6)$$

In the above formulas, $v$ denotes the 3D template model equal to a $3 \times 6449$ matrix. $I$ is a $1 \times 6449$ matrix with value one. $p$ is a $1 \times 3$ matrix with the value zero. $R_x$, $R_y$, and $R_z$ are defined with the following expressions:

$$R_z(\theta_z) = \begin{bmatrix} \cos(\theta_z) & -\sin(\theta_z) & 0 \\ \sin(\theta_z) & \cos(\theta_z) & 0 \\ 0 & 0 & 1 \end{bmatrix} \quad (7)$$

$$R_x(\theta_x) = \begin{bmatrix} 1 & 0 & 0 \\ 0 & \cos(\theta_x) & -\sin(\theta_x) \\ 0 & \sin(\theta_x) & \cos(\theta_x) \end{bmatrix} \quad (8)$$

$$R_y(\theta_y) = \begin{bmatrix} \cos(\theta_y) & 0 & \sin(\theta_y) \\ 0 & 1 & 0 \\ -\sin(\theta_y) & 0 & \cos(\theta_y) \end{bmatrix} \quad (9)$$

Where $\theta_x$, $\theta_y$ and $\theta_z$ are defined by the following formula:

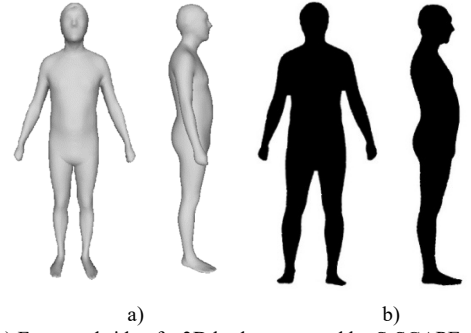

**Fig. 5:** (a) Front and side of a 3D body generated by S-SCAPE model (b) projection of the 3D body into the 2D space.

$$\theta_z = \frac{D_z \times \pi}{180} \quad (10)$$
$$\theta_x = \frac{D_x \times \pi}{180} \quad (11)$$
$$\theta_y = \frac{D_y \times \pi}{180} \quad (12)$$

$D_x$, $D_y$ and $D_z$ denote the projection degrees based on the x, y, and z-axis. To front projection $D_x$ is plugged to 90 and $D_y$ and $D_z$ is plugged to zero. To side projection $D_x$ and $D_y$ is plugged to 90 and $D_z$ is plugged to zero. The example of projection is illustrated in Fig. 5.

**Silhouette boundary extraction from 3D body:** In order to extract silhouette boundaries from 2D shapes resulting from the previous stage, the Moore-neighbor tracing algorithm is used again. Both silhouette boundaries from the front and side projection of the 3D body model and silhouette boundaries of the front and side of the person's photos are aligned with 2D rigid registration [10] and match using pairwise matching in the following stage.

### 3.3. Matching boundaries

The algorithm chosen for matching 3D models with 2D images depends on the features used for matching them. Different features from images and 3D models are extracted and used to compare, including feature points (minimum and maximum local points on silhouettes) [7],[8],[40], measures (chest girth, waist girth, etc.) [9], or combination of 2D key points (or 2D joints) and silhouettes (or masks) [38],[39],[35],[37],[34]. In this work, we exploit silhouette boundaries or contour as a comparison feature, and we use the 2D rigid registration algorithm to align and pairwise matching to match two silhouette boundaries.

**2D rigid registration:** This section briefly reviews the 2D rigid registration model introduced by Bouaziz et al. [10]. Let's consider silhouette's boundary of subject $Z$ and silhouette's boundary of projected 3D mesh $X$. In order to measure the distance between two boundaries, the source boundary $Z$ should be aligned to the target boundary $X$ by finding transformation $T$ that brings $Z$ in alignment with $X$. It can be defined as an energy minimization according to the following expressions:

$$arg_T \min E_{reg}(T, Z, X) \quad (13)$$

$$E_{reg} = E_{match} + E_{prior} \quad (14)$$

Where $E_{match}$ is alignment error and $E_{prior}$ denotes global rigidity for rigid objects like silhouette boundaries that limit the allowed transformations to rotations and translations. $E_{match}$ is defined according to the following expressions:

$$E_{match}(Z) = \sum_{i=1}^{n} \|z_i - P_x(z_i)\|_2^2 \quad (15)$$

Where $P_x(z_i)\ \mathbb{R}^2 \rightarrow \mathbb{R}^2$ using Euclidian distance returns the closest point on boundary $X$ from $z_i$. Kd-tree[11] is created to find the nearest point using a KNN search to find the closest point. $E_{prior}$ for rigid objects defined as a global rigidity that can be measured as:

$$E_{rigid}(Z, R, t) = \sum_{i=1}^{n} \|z_i - (Rx_i + t)\|_2^2 \quad (16)$$

Where $R\ \epsilon\ \mathbb{R}^{2\times 2}$ is a rotation matrix, and $t\ \epsilon\ \mathbb{R}^2$ is a translation vector. In this case, boundary $Z$ tries to follow a rigid transformation of boundary $X$. Fig. 6 (a) shows the alignment result using the 2D rigid registration algorithm of two silhouette boundaries.

**Pairwise matching:** After the alignment of two boundaries by energy minimization through Equation (14), pairwise matching is applied to obtain the matching error described in Equation (15). Fig. 6 (b) is an illustration of Pairwise matching resulting in a matching error value. Pairwise matching is calculated using KNN search on created Kd-tree. Each matching error value indicates the cost of its 24 model parameters.

**Weighting coefficients for body parts:** The matching error distance between some body parts is more important than the others. For example, the waist, chest, and height have a high priority for matching errors. We also assign a weight for each view. Our experiment indicated that defining a higher coefficient for side view than front view for fat body styles causes better results. It is noticeable that the matching error of the height of two silhouettes has the highest importance. Therefore, the distance between the highest and lowest point of two boundaries has the maximum weighting coefficient. Fig. 6 (c) illustrates the difference between the two silhouette's lowest and highest points. The weighting coefficient used in our system is listed in Table 2. Fig. 7 shows body mesh segmentation

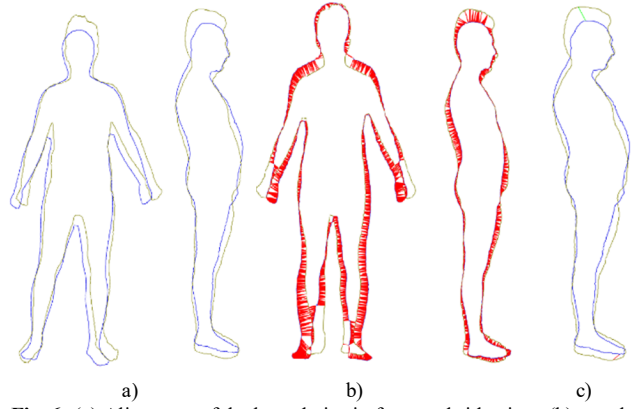

**Fig. 6:** (a) Alignment of the boundaries in front and side view. (b) matching error of silhouettes in front and side view shows with red lines (c) the red line is the difference between the lowest points, and the green line is the difference between the highest points of two silhouettes.

**Table 2:** weighting coefficients for different body parts to measure overall matching distance error.

| Body Parts | Weighting coefficient ($w_k$) |
|---|---|
| head | 2 |
| chest | 5 |
| waist | 5 |
| hip | 5 |
| leg | 2 |
| foot | 1 |
| arm | 3 |
| elbow | 2 |
| hand | 1 |
| height | 5 |
| The highest point | 5 |
| The lowest point | 5 |
| Front view | 2 |
| Side view | 3 |

of S-SCAPE model meshes. We separate high and low-importance body parts in this figure.

The result of projection by equations (4),(5), and (6) is a 2D points contour for each viewpoint. But due to mesh segmentation, we still know which 2D points belong to which body part. This is a key to finding body segmentation in 2D silhouettes. We normalize the 2D points of silhouette and the 2D point of the projected 3D mesh. Then using 2D rigid registration, we align them and match by pairwise matching. Therefore, we can connect each correspondent 2D point of individual body parts. Fig. 8 shows this matching for both views.

### 3.4. Body parameters optimization

In our system, 24 model parameters, including 20 shape parameters and four pose parameters, contribute to forming a 3D model using the statistical model. We need a parameter optimization algorithm to find the best parameters that match the silhouette to the 3D model. Different population-based methods of evolutionary algorithms like PSO and genetic algorithms can be used. We selected the genetic algorithm for the body parameter optimization. Therefore, every 24 parameters play the role of a chromosome in our algorithm. Moreover, each chromosome receives a cost or fitness value calculated by matching error of silhouettes.

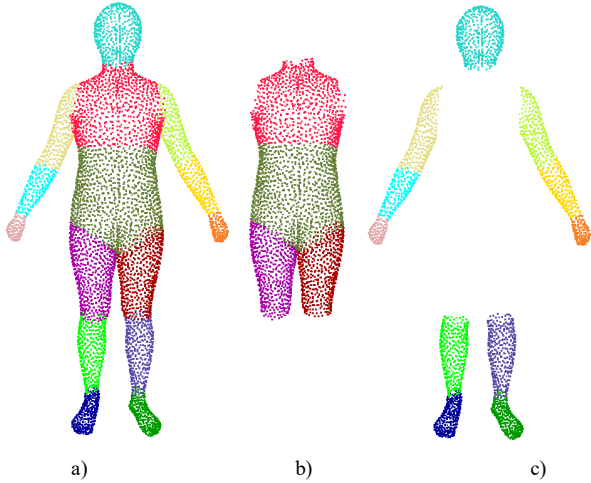

**Fig. 7:** (a) Body mesh segmentation defined for the S-SCAPE model. (b) body parts with high coefficients (c) body parts with low coefficients.

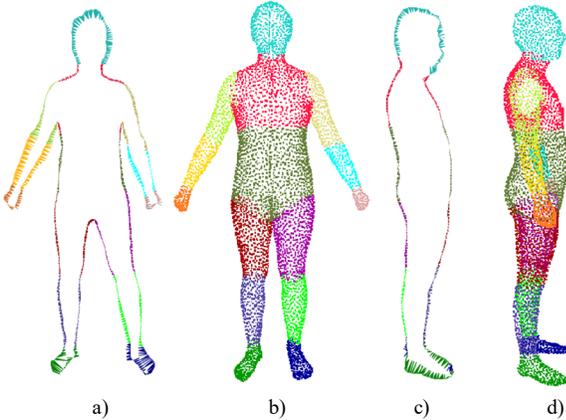

**Fig. 8:** (a) front (c) side view of Pairwise matching 2D points of contour to 2D points of projected mesh with body segmentation. (b) front (d) side view of 3D body segmentation, front view and side view.

**Objective function:** The objective function or fitness function in our optimization algorithm can be calculated by adding up the matching errors illustrated in Fig. 6 (b) and expressed by Equation (15). However, the weighting coefficients for different body parts defined in Table 2 should be applied to the distance error of each part shown in Fig. 8 (a,c) to emphasize what body parts are more important and should have more accuracy in parameter optimization. It is noticeable that the fitness value for each 3D model is calculated by the sum of the matching error of both the front and side views. Hence, the algorithm is restricted to control matching errors for both views simultaneously. Accordingly, the fitness function is calculated by:

$$f = w_{front}f_{front} + w_{side}f_{side} \quad (17)$$

Where $w_{front}$ and $w_{side}$ indicates the weighting coefficient defined in Table 2 and $f_{front}$ and $f_{side}$ are the matching error of each view and can be measured as:

$$f_{view} = \frac{1}{n}\left(\sum_{i=1}^{n} dis(z_i,x_i) \times w_k\right) \\ + \left(dis(z_{top},x_{top}) \times w_k\right) \\ + \left(dis(z_{down},x_{down}) \times w_k\right) \quad (18)$$

Where $dis(z_i,x_i)$ is the distance between 2 points $z$ and $x$. The sigma is equal to $E_{match}$ in Equation (15). $w_k$ is weighting coefficient according to Table 2. $z_{top}$ and $x_{top}$ are the highest points of two silhouettes and $z_{down}$ and $x_{down}$ are the lowest points of the two silhouettes demonstrated in Fig. 6 (c).

**Population Initialization:** In order to generate the first generation of the chromosomes, in other words, 24 shape and pose parameters for the S-SCAPE model, random numbers are generated in a specific range. The range for the first 20 of 24 parameters that are shape parameters should be between -3 and +3. These 20 generated numbers are multiplied by the square root of the corresponding 20 top eigenvalues of PCs to create $\pm 3 \times \sigma$ ($\pm 3 \times \text{Std. Dev}$) coefficients, which is equal to parameter $\beta$ in Equation (1).

In order to handle the slight variation in the angles of the arms and legs of the person, the 30 DOF for joint angle was defined. So, the last four of the 24 parameters, which are pose parameters, should be a random number between 0 and 30 degrees. Generally, we consider 30 chromosomes for each generation, and each chromosome is a 24 random number. Therefore we define $30 \times 24$ random parameters for population initialization.

**Selection, Crossover, and Mutation:** After each iteration, chromosomes are sorted by fitness values defined as Equations (17). Then one-third of them, or ten, which are the lowest fitness values, are removed. One-point and two-point crossover randomly apply on the remind chromosome to generate the next 30 chromosomes.

In general, optimization-solving algorithms converge to a local minimum. Therefore, it would be inevitable for the pop-ulation of chromosomes to become too similar. In such a situation, it is necessary to provide diversity in generated 3D bodies by changing parameters randomly. Therefore, a mutation is randomly applied to three of the 30 chromosomes. In these three randomly chosen chromosomes, the value of 5 genes (from 24 genes) is randomly changed by mutation values.

## 4. Validation and results

In this section, we report the results of the quantitative and qualitative experiments we have performed on our method.

### 4.1. Quantitative Experiments

To quantitatively evaluate our approach, like previous works [4],[12],[15],[21],[32],[14], we run our experiment on synthetic data. The same 16 anthropometric body measurements illustrated in Fig. 10 are compared in our system with other recent works. CAESAR dataset, which used in some works [12][32], is commercial, and due to the lack of access to it in our work, we generate 500 body meshes as synthetic data by normalized S-SCAPE model, including 250 males and 250 females. Our synthetic data is 500 body meshes generated by the normalized S-SCAPE model [1], including 250 males and 250 females.

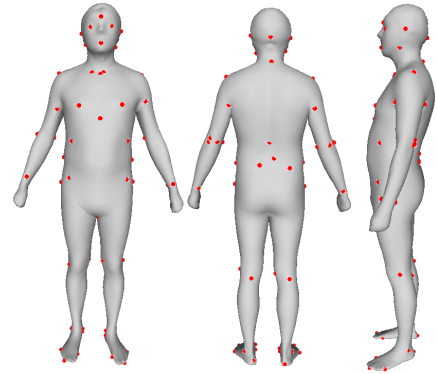

**Fig. 9:** 73 landmarks of CAESAR dataset used for extract 16 anthropometric measurements.

**Table 3:** Error comparisons on 16 anthropometric measurements of the ground truth and the estimated meshes in the proposed method compared with similar recent works. Errors are represented as Mean ± Std. Dev and are expressed in millimeters.

| Measurements | Proposed Method | BfSNet Smith et al. [4] | HS-2-Net-MM Dibra et al. [21] | UF-US-HKS-2 Dibra et al. [32] | Boisvert et al. [12] | Chen et al. [55][15] | Xi et al. [14] |
|---|---|---|---|---|---|---|---|
| A.Head circumference | 3.1±3.5 | 5.1±6.4 | **2**±3 | 3.2±2.6 | 10±12 | 23±27 | 50±60 |
| B.Neck circumference | 2.1±2 | 3.0±3.9 | 2±1 | **1.9**±1.5 | 11±13 | 27±34 | 59±72 |
| C.Shoulder to crotch length | 2.2±3.5 | **1.5**±2.2 | 3±5 | 4.2±3.4 | 4±5 | 52±65 | 119±150 |
| D.Chest circumference | 2.5±3.2 | 4.7±7.7 | **2**±1 | 5.6±4.7 | 10±12 | 18±22 | 36±45 |
| E.Waist circumference | 5.1±7.1 | **4.8**±7.5 | 7±5 | 7.1±5.8 | 22±23 | 37±39 | 55±62 |
| F.Pelvis circumference | 3.3±5.5 | **3.0**±5.1 | 4±4 | 6.9±5.6 | 11±12 | 15±19 | 23±28 |
| G.Wrist circumference | 3±3.1 | 2.5±3.3 | 2±2 | **1.6**±1.3 | 9±12 | 24±30 | 56±70 |
| H.Bicep circumference | 3.2±3.4 | 2.7±3.8 | **2**±1 | 2.6±2.1 | 17±22 | 59±76 | 146±177 |
| I.Forearm circumference | 3.1±3.2 | 1.9±2.5 | **1**±1 | 2.2±1.9 | 16±20 | 76±100 | 182±230 |
| J.Arm length | 3.9±2.8 | **1.7**±2.4 | 3±2 | 2.3±1.9 | 15±21 | 53±73 | 109±141 |
| K.Inside leg length | 3.2±4.8 | **1.5**±2.7 | 9±6 | 4.3±3.8 | 6±7 | 9±12 | 19±24 |
| L.Thigh circumference | 2.6±4.4 | **2.4**±4.0 | 6±4 | 5.1±4.3 | 9±12 | 19±25 | 35±44 |
| M.Calf circumference | 3.2±2.1 | **2.3**±3.6 | 3±1 | 2.7±1.9 | 6±7 | 16±21 | 33±42 |
| N.Ankle circumference | 2.9±1.8 | 2.1±2.8 | 2±1 | **1.4**±1.1 | 14±16 | 28±35 | 61±78 |
| O.Overall height | 5.1±6.2 | **2.3**±4.6 | 12±10 | 7.1±5.5 | 9±12 | 21±27 | 49±62 |
| P. Shoulder breadth | 2.1±2.9 | **1.9**±2.5 | 2±4 | 2.1±1.8 | 6±7 | 12±15 | 24±31 |

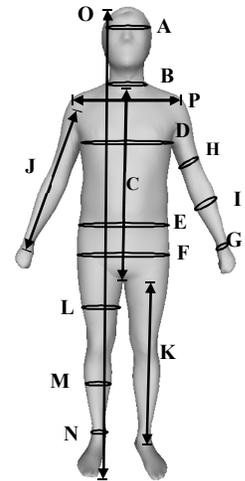

**Fig. 10:** 16 body measurements.

**Table 4:** Error comparisons on visually important body parts including seven anthropometric measurements of the ground truth and the estimated meshes in the proposed method with similar recent works. Errors are represented as Mean ± Std. Dev and are expressed in millimeters.

| Measurements | Proposed Method | BfSNet Smith et al. [4] | HS-2-Net-MM Dibra et al. [21] | UF-US-HKS-2 Dibra et al. [32] | Boisvert et al. [12] | Chen et al. [15] | Xi et al. [14] |
|---|---|---|---|---|---|---|---|
| B.Neck circumference | 2.1±2 | 3.0±3.9 | 2±1 | 1.9±1.5 | 11±13 | 27±34 | 59±72 |
| C.Shoulder to crotch length | 2.2±3.5 | **1.5**±2.2 | 3±5 | 4.2±3.4 | 4±5 | 52±65 | 119±150 |
| D.Chest circumference | 2.5±3.2 | 4.7±7.7 | 2±1 | 5.6±4.7 | 10±12 | 18±22 | 36±45 |
| E.Waist circumference | 5.1±7.1 | **4.8**±7.5 | 7±5 | 7.1±5.8 | 22±23 | 37±39 | 55±62 |
| F.Pelvis circumference | 3.3±5.5 | **3.0**±5.1 | 4±4 | 6.9±5.6 | 11±12 | 15±19 | 23±28 |
| L.Thigh circumference | 2.6±4.4 | **2.4**±4.0 | 6±4 | 5.1±4.3 | 9±12 | 19±25 | 35±44 |
| P. Shoulder breadth | 2.1±2.9 | **1.9**±2.5 | 2±4 | 2.1±1.8 | 6±7 | 12±15 | 24±31 |
| Mean measurement error | **2.8±4 mm** | 3±4.7 | 3.7±3.4 mm | 4.7±4.3 mm | 10.4±12 | 25.7±31.29 | 50.1±61.7 |

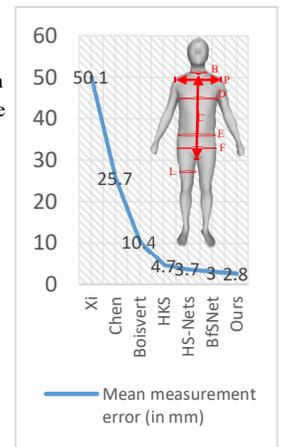

**Fig. 11:** Seven visually important body measurements.

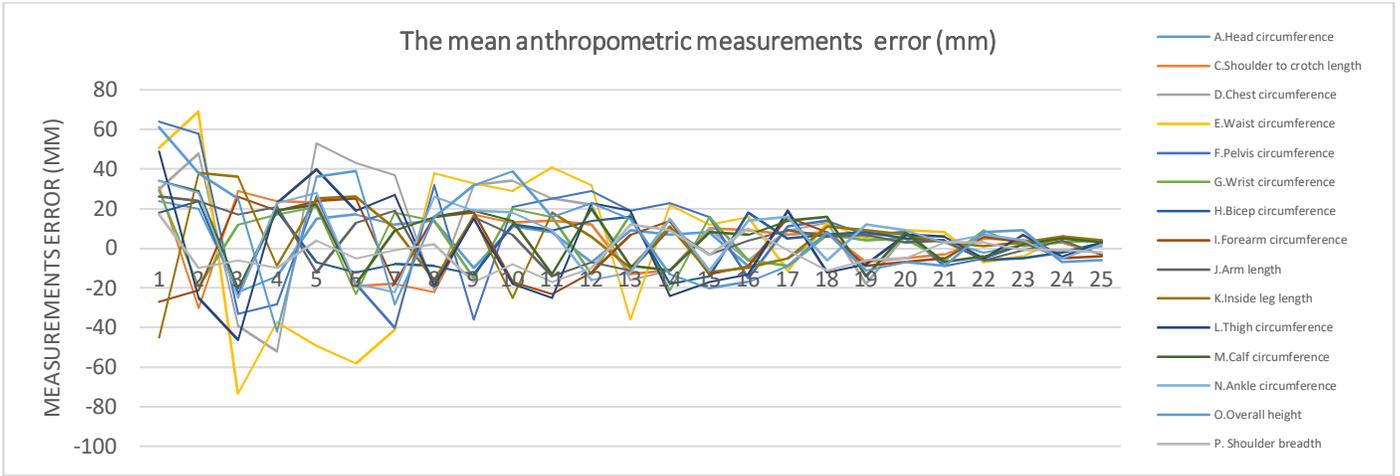

**Fig. 12:** the mean measurement errors (in millimeters) of tests after 25 iteration.

In order to find exact 3D points for calculating anthropometric measurements in synthetic 3D mesh, we used 73 CAESAR anthropometric landmarks depicted in Fig. 9. In 16 measurements shown in Fig. 10, straight lines represent the Euclidean distance of width or height, and ellipses illustrate the geodesic distance of circumferences. Geodesic distance is measured on the body surface. To scale these measures, we used body height. Similar works [32],[54],[4] to scale anthropometric measurements extracted from images or silhouettes used body height.

We calculate the difference between the estimated values and ground truth values for all 16 measurements. Table 3 shows the mean error and standard deviation results for all the test meshes. The data shows that the accuracy of our proposed method in these seven measurements of important body parts is very close to the best values of recent works, especially for neck, chest, waist circumference, and shoulder breadth. We analysis separately seven measures related to visually important body parts illustrated in Fig. 11. Table 4 shows that the proposed method has a more accurate result than other methods in these seven crucial measures, with a mean measurement error of 2.8 mm.

To observe the parameters optimization progress, we collected and averaged the measurement errors of tests in each iteration. Fig. 12 illustrates a meaningful graph that shows parameters optimization progress during 25 iterations that become close to zero at the end.

### 4.2. Qualitative Experiments

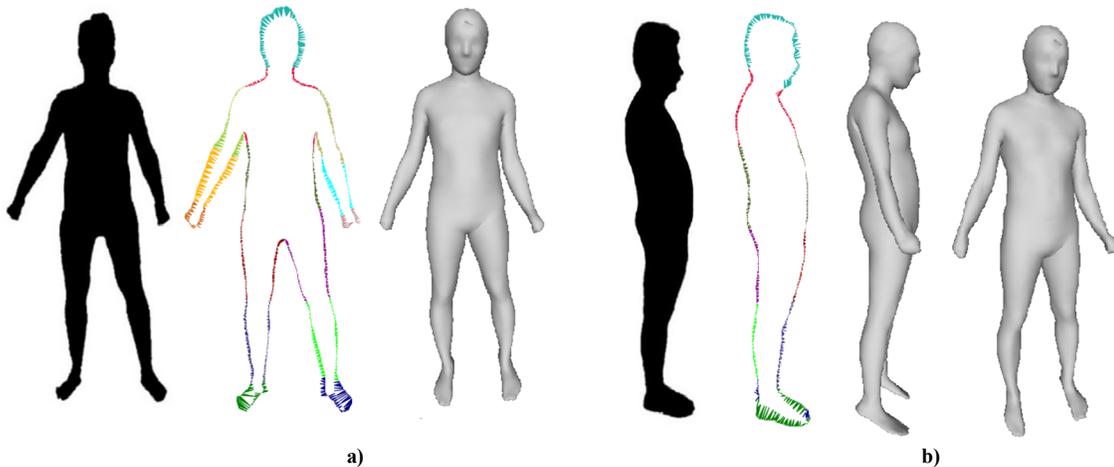

**Fig. 13:** Qualitative result of proposed method on a front and side silhouettes from real person. **first image**: silhouettes, **second image**: matching contours, **third image**: body model result. (a) front view. (b) side view.

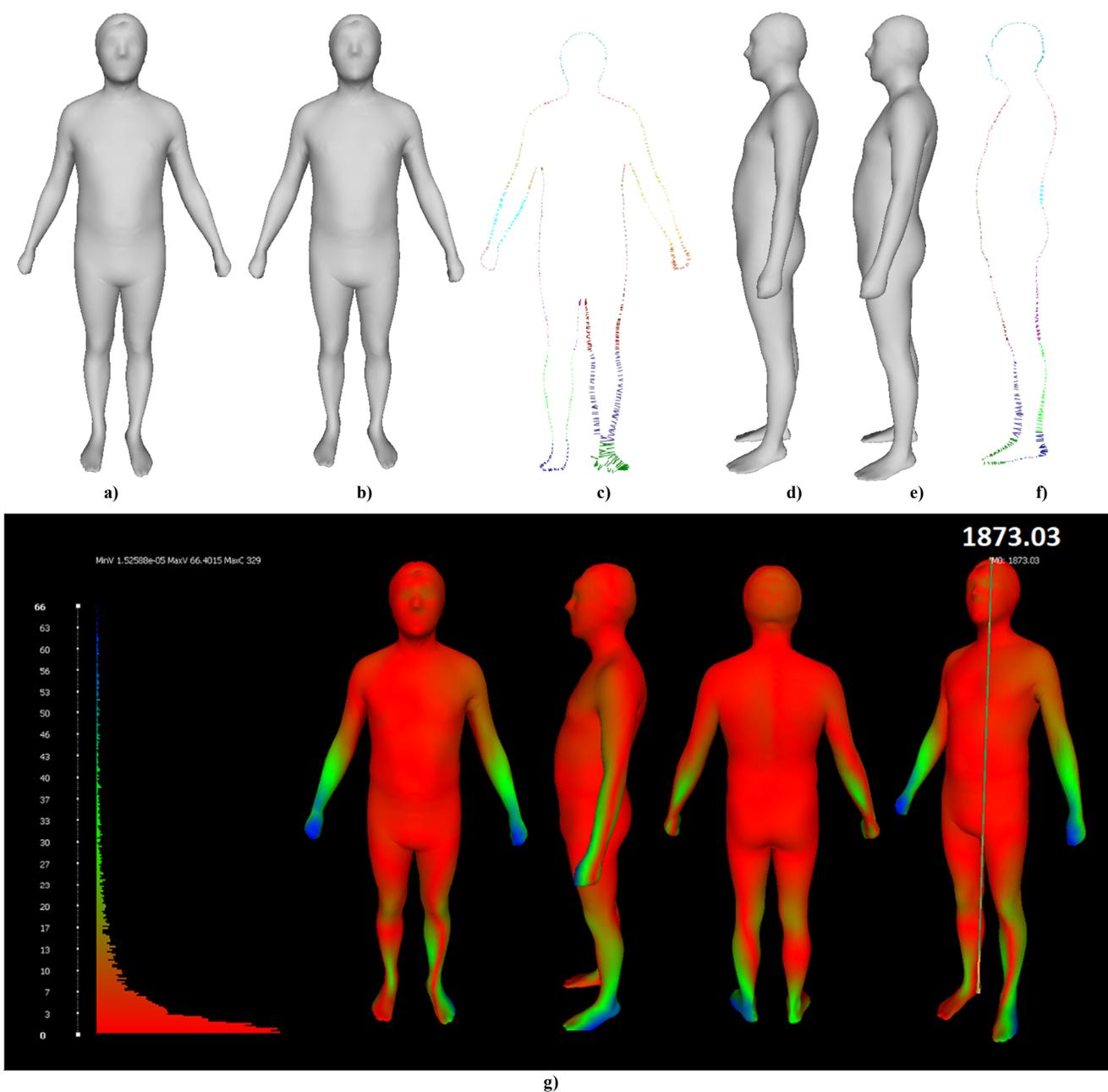

**Fig. 14:** Qualitative result of the proposed method on front and side silhouettes extracted from a ground truth mesh of the S-SCAPE dataset. (a) estimated mesh, front view. (b) ground truth mesh, front view. (c) matching contours, front view (d) estimated mesh, side view. (e) ground truth mesh, side view. (f) matching contour, side view. (g) compute Hausdorff distance of estimated mesh and ground truth mesh using Meshlab software.

For the first experiment, we demonstrate 3D body results of the front and side silhouettes extracted from real person images in Fig. 13. It is noticeable to remind that before starting the reconstruction process, we undistort the captured images using the camera calibrator discussed in the photography assumptions section.

We also performed experiments on meshes from two different datasets to compute the exact difference between estimated and grand truth meshes. So, for the second experiment, we chose a synthetic mesh from the S-SCAPE dataset (6449 vertices per mesh) to reconstruct the 3D mesh from its silhouettes. The Hausdorff distance of estimated mesh and ground truth mesh was computed using Meshlab software and is shown in Fig. 14. To define our unit of measurement in Meshlab, we measured the height of the body mesh as a scale for distances. According to this measure, we noticed that the unit of measurement for this specific mesh type could be considered a millimeter. In the third experiment, we chose a synthetic mesh from the CAESAR dataset (10777 vertices per mesh) as a ground truth mesh. To compare the estimated mesh with the ground truth mesh, we rescaled the chosen CAESAR mesh to the S-SCAPE scale by multiplying its scale by 1000 for X, Y, and Z and -900 for Z translation, and -90 for Z rotation. The result of this experiment is illustrated in Fig. 15.

In the fourth qualitative experiment, we evaluated our method on three distinct body styles, including normal, fat and thin. The result of the reconstruction is shown in Fig. 16. The result shows that the method performance does not differ in various body styles. For the sixth and last experiments, we run the same previous test but with equal weighting coefficients for all body parts. The result is depicted in Fig. 17. The comparison of Fig. 17 with Fig. 16 demonstrates how important these coefficients are in the accuracy of the proposed method.

## 5. Discussion and future works

In comparison with previous works, the proposed method has both advantages and disadvantages. The first benefit is that it does not require training data; consequently, there is no time for training. At the same time, the algorithm running time is not optimum. Each iteration takes around six seconds. This running time is for performing the proposed method on Intel Core™ i7(4800MQ). In simple geometry bodies, the method reaches the best parameters in early iterations, 8 to 14, and it takes about 20 to 30 iterations for detailed cases. In similar work, Dibra et al.[21] training time is about 30 minutes per epoch, and the running time for the full algorithm is 0.45 seconds. The next proposed method by Dibra et al. [32] takes about 50 min per epoch to train, and the running time takes about 0.3 seconds. The method proposed by Boisvert et al. [12] takes 3 minutes and 36 seconds to process. Chen et al.

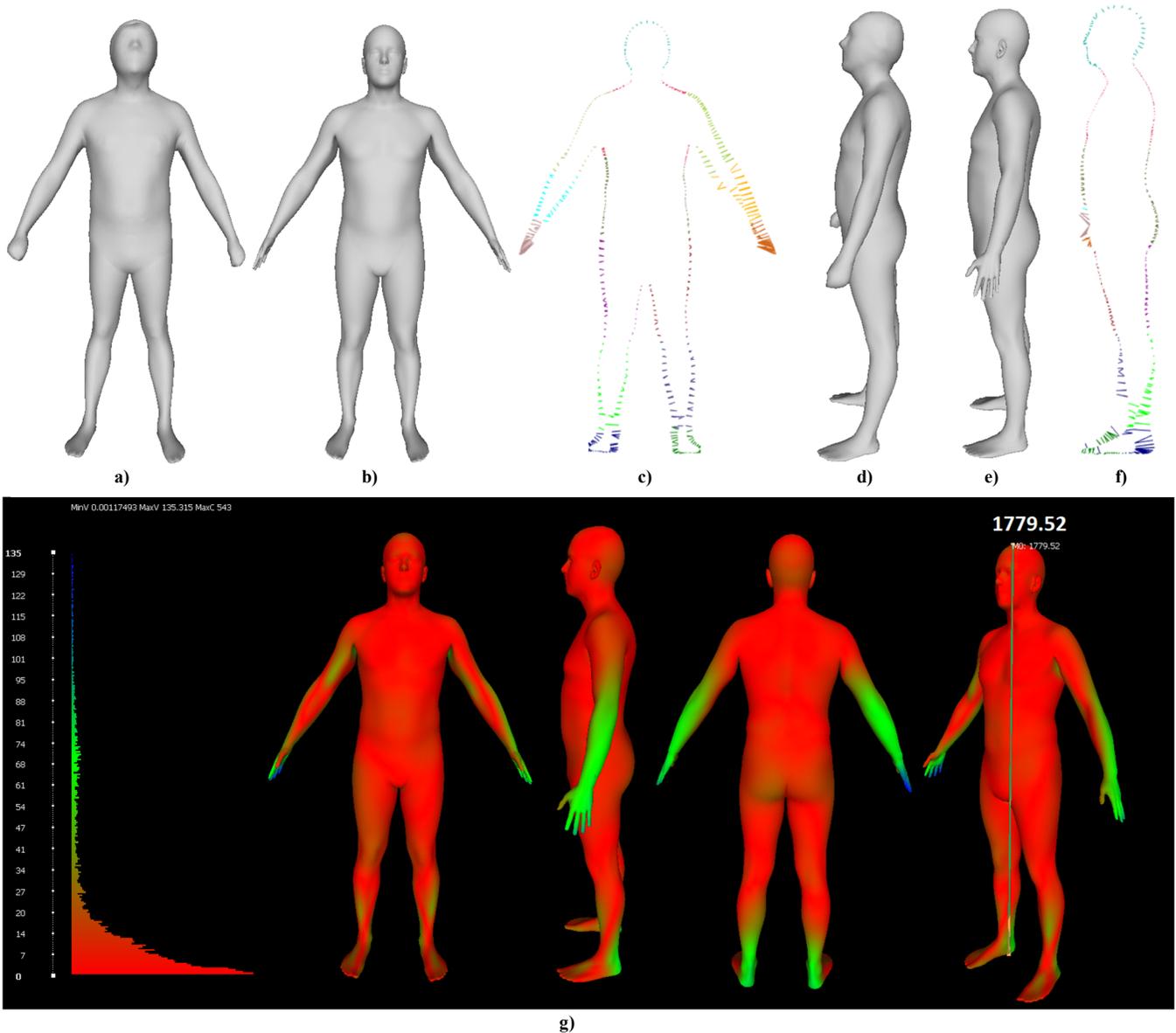

**Fig. 15:** Qualitative result of the proposed method on front and side silhouettes extracted from a ground truth mesh of the CAESAR dataset. (a) estimated mesh, front view. (b) ground truth mesh, front view. (c) matching contours, front view (d) estimated mesh, side view. (e) ground truth mesh, side view. (f) matching contour, side view. (g) compute Hausdorff distance of estimated mesh and ground truth mesh using Meshlab software.

[55] For each image, they reached a running time of 10 - 15 minutes to predict ten candidates.

We limited our shape space to the first 20 PCs like Dibra et al. [32], which may slightly diminish the algorithm's efficiency in reconstructing a few body shapes with more details, especially in fat body styles. Although increasing the PCs exploited in the statistical model can improve the accuracy, it raises the processing time of each algorithm iteration. Furthermore, local shape deformation used by some works [12],[54] to fit the model to silhouette can be used in future works to improve body shape similarity as a final stage.

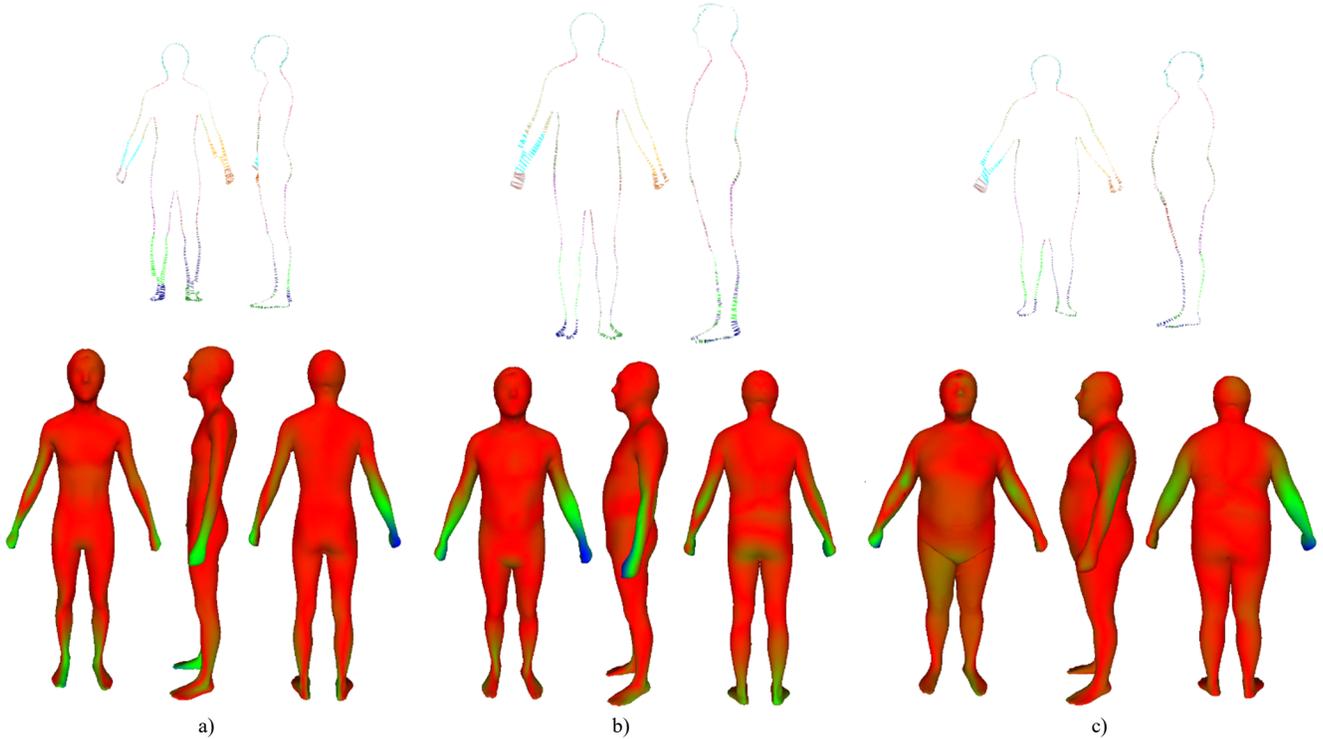

**Fig. 16:** Comparison of ground truth meshes and estimated meshes in different body styles, including fat, thin, and normal. Red and yellow color shows the slightest difference and green, blue and black indicate the biggest difference respectively. **top image:** matching contours. **bottom image:** matching meshes. (a) normal (b) thin (c) fat model.

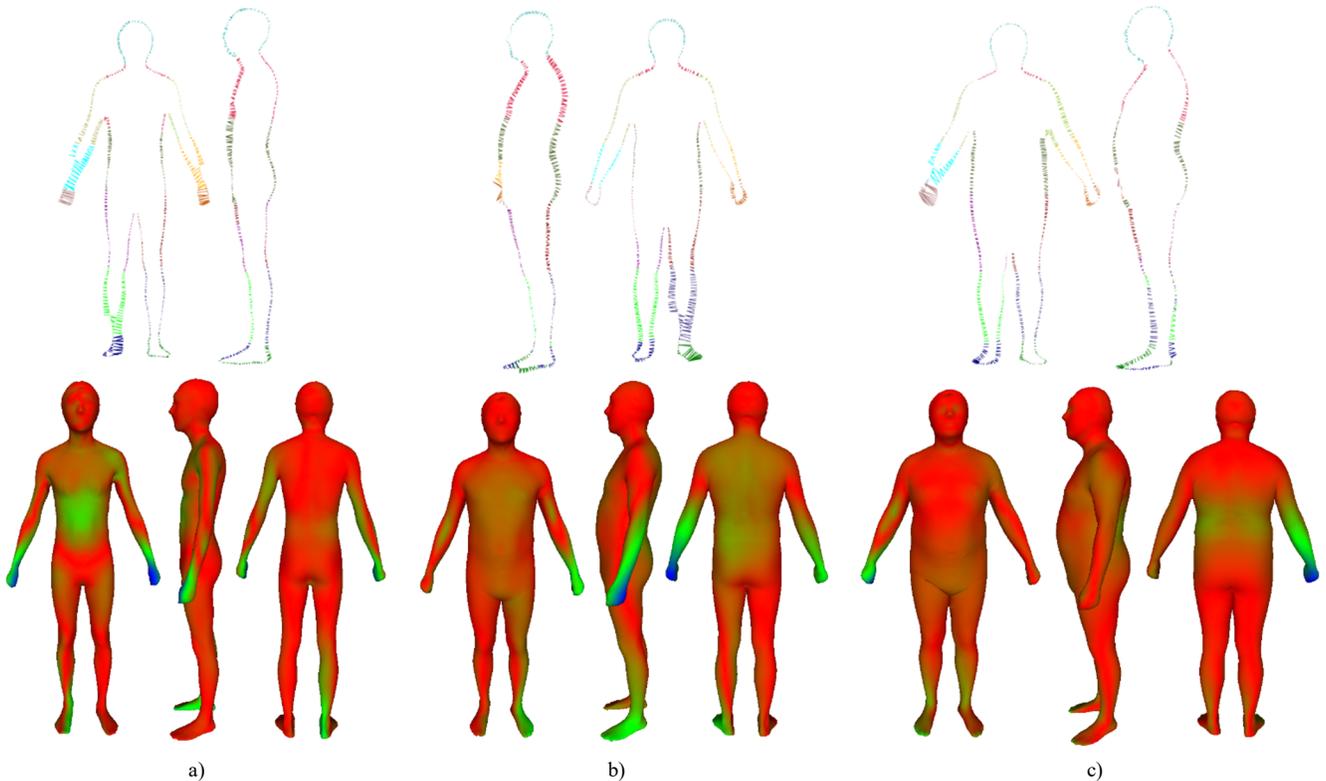

**Fig. 17:** Results with low accuracy if weighting coefficient of all body parts are same. Comparison of ground truth meshes and estimated meshes in different body styles, including fat, thin, and normal. Red and yellow color shows the slightest difference and green, blue and black indicate the biggest difference. respectively. (a) normal (b) thin (c) fat model.

## 6. Conclusion

This paper proposes novel adjustable algorithms for reconstructing 3D body shapes from front and side silhouettes. Other approaches train a deep network by silhouettes and key points. However, they cannot predict shape parameters that precisely fit the model to the body contour, especially in the torso, which is visually important. In contrast, in the proposed algorithm, we can enhance the accuracy of more important body parts than others. To optimize shape parameters, the algorithm in each iteration minimizes the objective function with multiple parameters, including distances and coefficients for body parts, the highest and lowest point, and front and side view. To match correspondence body parts in projected 3D mesh and 2D silhouette, we exploit body segmentation of meshes. The target person can be slightly flexible in the arms and legs angle.

## Declaration

All authors declare that they have no conflicts of interest. All authors certify that they have NO affiliations with or involvement in any organization or entity with any financial interest.